\documentclass{llncs}

\usepackage{amsmath}
\usepackage{latexsym}
\usepackage{epsfig}

\begin{document}

\bibliographystyle{plain}

\pagestyle{headings}

\mainmatter

\title{Reduced cost-based ranking for generating promising subproblems}


\author{M. Milano\inst{1} and W.J. van Hoeve\inst{2}}

\institute{DEIS, University of Bologna, Viale Risorgimento 2, 40136 Bologna,
Italy\\
\email{mmilano@deis.unibo.it}\\
\texttt{http://www-lia.deis.unibo.it/Staff/MichelaMilano/} \and
CWI, P.O. Box 94079, 1090 GB Amsterdam, The Netherlands\\
\email{w.j.van.hoeve@cwi.nl}\\
\texttt{http://www.cwi.nl/\~{ }wjvh/}}

\maketitle

\begin{abstract}
In this paper, we propose an effective search procedure that
interleaves two steps: subproblem generation and subproblem
solution. We mainly focus on the first part. It consists of a
variable domain value ranking based on reduced costs. Exploiting
the ranking, we generate, in a Limited Discrepancy Search tree,
the most promising subproblems first. An interesting result is
that reduced costs provide a very precise ranking that allows to
almost always find the optimal solution in the first generated
subproblem, even if its dimension is significantly smaller than
that of the original problem. Concerning the proof of optimality,
we exploit a way to increase the lower bound for subproblems at
higher discrepancies. We show experimental results on the TSP and
its time constrained variant to show the effectiveness of the
proposed approach, but the technique could be generalized for
other problems.
\end{abstract}

\section{Introduction}
In recent years, combinatorial optimization problems have been
tackled with hybrid methods and/or hybrid solvers
\cite{CP-AI-OR99,CP-AI-OR00,CP-AI-OR01,CP-AI-OR02}. The use of
problem relaxations, decomposition, cutting planes generation
techniques in a Constraint Programming (CP) framework are only
some examples. Many hybrid approaches are based on the use of a
relaxation $R$, i.e. an easier problem derived from the original
one by removing (or relaxing) some constraints. Solving $R$ to
optimality provides a bound on the original problem. Moreover,
when the relaxation is a linear problem, we can derive {\em
reduced costs} through dual variables often with no additional
computational cost. Reduced costs provide an optimistic esteem (a
bound) of each variable-value assignment cost. These results have
been successfully used for pruning the search space and for
guiding the search toward promising regions (see
\cite{NostroCP99}) in many applications like TSP
\cite{NostroCP98}, TSPTW \cite{FLM2002}, scheduling with sequence
dependent setup times \cite{SchedSetUpAIPS} and multimedia
applications \cite{FahleSellmanCPAIOR_small}.

\begin{figure}[t]
\begin{center}
\epsfig{figure=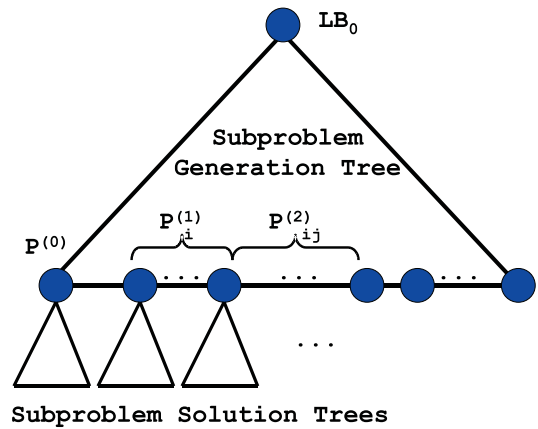,width=0.60\textwidth}
\end{center}
\caption{The structure of the search tree.} \label{fig:tree}
\end{figure}

We propose here a solution method, depicted in Figure \ref{fig:tree}, based
on a two step search procedure that interleaves ($i$) subproblem generation
and ($ii$) subproblem solution. In detail, we solve a relaxation of the
problem at the root node and we use reduced costs to rank domain values;
then we partition the domain of each variable $X_i$ in two sets, i.e., the
{\em good} part $D_i^{good}$ and the {\em bad} part $D_i^{bad}$. We search
the tree generated by using a strategy imposing on the left branch the
branching constraint $X_i \in D_i^{good}$ while on the right branch we
impose $X_i \in D_i^{bad}$. At each leaf of the {\em subproblem generation
tree}, we have a subproblem which can now be solved (in the {\em subproblem
solution tree}).

Exploring with a Limited Discrepancy Strategy the resulting search space,
we obtain that the first generated subproblems are supposed to be the most
promising and are likely to contain the optimal solution. In fact, if the
ranking criterion is effective (as the experimental results will show), the
first generated subproblem (discrepancy equal to 0) $P^{(0)}$, where all
variables range on the good domain part, is likely to contain the optimal
solution. The following generated subproblems (discrepancy equal to 1)
$P^{(1)}_i$ have all variables but the $i$-th ranging on the good domain
and are likely to contain worse solutions with respect to $P^{(0)}$, but
still good. Clearly, subproblems at higher discrepancies are supposed to
contain the worst solutions.

A surprising aspect of this method is that even by using low cardinality
good sets, {\bf we almost always find the optimal solution in the first
generated subproblem}. Thus, reduced costs provide extremely useful
information indicating for each variable which values are the most
promising. Moreover, this property of reduced costs is independent of the
tightness of the relaxation. Tight relaxations are essential for the proof
of optimality, but not for the quality of reduced costs. Solving only the
first subproblem, we obtain a very effective incomplete method that finds
the optimal solution in almost all test instances.

To be complete, the method should solve all subproblems for all
discrepancies to prove optimality. Clearly, even if each subproblem could
be efficiently solved, if all of them should be considered, the proposed
approach would not be applicable. The idea is that by generating the
optimal solution soon and tightening the lower bound with considerations
based on the discrepancies shown in the paper, we do not have to explore
all subproblems, but we can prune many of them.

In this paper, we have considered as an example the Travelling Salesman
Problem and its time constrained variant, but the technique could be
applied to a large family of problems.

The contribution of this paper is twofold: ($i$) we show that
reduced costs provide an extremely precise indication for
generating promising subproblems, and ($ii$) we show that LDS can
be used to effectively order the subproblems. In addition, the use
of discrepancies enables to tighten the problem bounds for each
subproblem.

The paper is organized as follows: in Section~\ref{sc:preliminaries} we
give preliminaries on Limited Discrepancy Search (LDS), on the TSP and its
time constrained variant. In Section~\ref{sc:method} we describe the
proposed method in detail. Section~\ref{implementation} discusses the
implementation, focussing mainly on the generation of the subproblems using
LDS. The quality of the reduced cost-based ranking is considered in
Section~\ref{sc:quality}. In this section also the size of subproblems is
tuned. Section~\ref{sc:results} presents the computational results.
Conclusion and future work follow.

\section{Preliminaries}\label{sc:preliminaries}

\subsection{Limited Discrepancy Search}

Limited Discrepancy Search (LDS) was first introduced by Harvey
and Ginsberg \cite{harvey95}. The idea is that one can often find
the optimal solution by exploring only a small fraction of the
space by relying on tuned (often problem dependent) heuristics.
However, a perfect heuristic is not always available. LDS
addresses the problem of what to do when the heuristic fails.

Thus, at each node of the search tree, the heuristic is supposed
to provide the {\em good} choice (corresponding to the leftmost
branch) among possible alternative branches. Any other choice
would be {\em bad} and is called a {\em discrepancy}. In LDS, one
tries to find first the solution with as few discrepancies as
possible. In fact, a perfect heuristic would provide us the
optimal solution immediately. Since this is not often the case, we
have to increase the number of discrepancies so as to make it
possible to find the optimal solution after correcting the
mistakes made by the heuristic. However, the goal is to use only
few discrepancies since in general good solutions are provided
soon.

LDS builds a search tree in the following way: the first solution
explored is that suggested by the heuristic. Then solutions that
follow the heuristic for every variable but one are explored:
these solutions are that of discrepancy equal to one. Then,
solutions at discrepancy equal to two are explored and so on.

It has been shown that this search strategy achieves a significant
cutoff of the total number of nodes with respect to a depth first
search with chronological backtracking and iterative sampling
\cite{Langley92}.

\subsection{TSP and TSPTW}
\label{problem}

  Let $G=(V,A)$ be a digraph, where $V = \{1, \dots,
n\}$ is the vertex set and $A = \{(i,j) : i,j \in V \}$ the arc
set, and let $c_{ij} \ge 0$ be the cost associated with arc $(i,j)
\in A$ (with $c_{ii} = +\infty$ for each $i \in V$). A {\em
Hamiltonian Circuit} ({\em tour}) of $G$ is a partial digraph
$\bar G = (V,\bar A)$ of $G$ such that: $|\bar A| = n$ and for
each pair of distinct vertices $v_1,v_2 \in V$, both paths from
$v_1$ to $v_2$ and from $v_2$ to $v_1$ exist in $\bar G$ (i.e.
digraph $\bar G$ is {\em strongly connected}).

The Travelling Salesman Problem (TSP) looks for a Hamiltonian circuit $G^*
= (V,A^*)$ whose cost $\sum_{(i,j) \in A^*} c_{ij}$ is a minimum.

A classic Integer Linear Programming formulation for TSP is as
follows:

\begin{eqnarray}
\label{eq-1}
& v(TSP)           & = \min  \sum_{i\in V} \sum_{j\in V} c_{ij}\, x_{ij}\\
\label{eq-2}
& \hbox{subject~to} &  \sum_{i \in V} x_{ij} =  1,   \qquad j\in V\\
\label{eq-3}
&                   & \sum_{j \in V} x_{ij} =  1,   \qquad i\in V\\
\label{eq-4}
&                   & \sum_{i \in S} \sum_{j \in V \setminus S}
x_{ij}  \ge  1,
                                        \qquad S \subset V,\; S \not =
\emptyset \\
\label{eq-5}
&                   & x_{ij}  \hbox{~integer,~~}   i,j \in V
\end{eqnarray}

\noindent where $x_{ij} = 1$ if and only if arc $(i,j)$ is part of
the solution. Constraints (\ref{eq-2}) and (\ref{eq-3}) impose
in-degree and out-degree of each vertex equal to one, whereas
constraints (\ref{eq-4}) impose strong connectivity.

Constraint Programming relies in general on a different model where we have a
domain variable
${\rm\it Next}_i$ (resp. ${\rm\it Prev}_i$) that identifies cities visited
after
(resp. before)
node $i$. Domain variable $Cost_i$ identifies the cost to be paid to go from
node $i$ to node
${\rm\it Next}_i$.

Clearly, we need a mapping between the CP model and the ILP model:
${\rm\it Next}_{i} = j \; \Leftrightarrow \; x_{ij} = 1$. The
domain of variable ${\rm\it Next}_{i}$ will be denoted as $D_i$.
Initially, $D_i = \{1, \dots, n\}$.

The Travelling Salesman Problem with Time Windows (TSPTW) is a
time constrained variant of the TSP where the service at a node
$i$ should begin within a time window $[a_{i}, b_{i}]$ associated
to the node. Early arrivals are allowed, in the sense that the
vehicle can arrive before the time window lower bound. However, in
this case the vehicle has to wait until the node is ready for the
beginning of service.

As concerns the CP model for the TSPTW, we add to the TSP model a domain
variable ${\rm\it
Start}_i$ which identifies the time at which the service begins at node $i$.

A well known relaxation of the TSP and TSPTW obtained by
eliminating from the TSP model constraints (\ref{eq-4}) and time
windows constraints is the {\em Linear Assignment Problem} (AP)
(see \cite{DM97} for a survey). AP is the graph theory problem of
finding a set of {\em disjoint} subtours such that all the
vertices in $V$ are visited and the overall cost is a minimum.
When the digraph is complete, as in our case, AP always has an
optimal integer solution, and, if such solution is composed by a
single tour, is then optimal for TSP satisfying constraints
(\ref{eq-4}).

The information provided by the AP relaxation is a lower bound $LB$ for the
original problem and
the reduced cost matrix $\bar c$. At each node of the decision tree, each
$\bar
c_{ij}$ estimates
the additional cost to pay to put arc $(i,j)$ in the solution. More formally,
a
valid lower bound
for the problem where $x_{ij}=1$ is $LB|_{x_{ij}=1} = LB + \bar c_{ij}$. It is
well-known that when
the AP optimal solution is obtained through a {\em primal-dual} algorithm, as
in
our case (we use a
C++ adaptation of the AP code described in \cite{CMT88}), the reduced cost
values are obtained
without extra computational effort during the AP solution. The solution of the
AP relaxation at the
root node requires in the worst case $O(n^{3})$, whereas each following AP
solution can be
efficiently computed in $O(n^{2})$ time through a single augmenting path step
(see \cite{CMT88} for
details). However, the AP does not provide a tight bound neither for the TSP
nor
for the TSPTW.
Therefore we will improve the relaxation in Section~\ref{lagr}.

\section{The Proposed Method}\label{sc:method}
In this section we describe the method proposed in this paper. It
is based on two interleaved steps: subproblem generation and
subproblem solution. The first step is based on the optimal
solution of a (possibly tight) relaxation of the original problem.
The relaxation provides a lower bound for the original problem and
the reduced cost matrix. Reduced costs are used for ranking (the
lower the better) variable domain values. Each domain is now
partitioned according to this ranking in two sets called the {\em
good} set and the {\em bad} set. The cardinality of the good set
is problem dependent and is experimentally defined. However, it
should be significantly lower than the dimension of the original
domains.

Exploiting this ranking, the search proceeds by choosing at each
node the branching constraint that imposes the variable to range
on the good domain, while on backtracking we impose the variable
to range on the bad domain. By exploring the resulting search tree
by using an LDS strategy we generate first the most promising
problems, i.e., those where no or few variables range on the bad
sets.

Each time we generate a subproblem, the second step starts for optimally
solving
it. Experimental
results will show that, surprisingly, even if the subproblems are small, the
first generated
subproblem almost always contains the optimal solution. The proof of
optimality
should then proceed
by solving the remaining problems. Therefore, a tight initial lower bound is
essential. Moreover,
by using some considerations on discrepancies, we can increase the bound and
prove optimality fast.

The idea of ranking domain values has been previously used in
incomplete algorithms, like GRASP \cite{FR95}. The idea is to
produce for each variable the so called Restricted Candidate List
(RCL), and explore the subproblem generated only by RCLs for each
variable. This method provides in general a good starting point
for performing local search. Our ranking method could in principle
be applied to GRASP-like algorithms.

Another connection can be made with iterative broadening
\cite{iterative_broadening}, where one can
view the breadth cutoff as corresponding to the cardinality of our good sets.
The first generated
subproblem of both approaches is then the same. However, iterative broadening
behaves differently
on backtracking (it gradually restarts increasing the breadth cutoff).

\subsection{Linear relaxation}
\label{lagr}

In Section \ref{problem} we presented a relaxation, the Linear
Assignment Problem (AP), for both the TSP and TSPTW. This
relaxation is indeed not very tight and does not provide a good
lower bound. We can improve it by adding cutting planes. Many
different kinds of cutting planes for these problems have been
proposed and the corresponding separation procedure has been
defined \cite{PR90}. In this paper, we used the Sub-tour
Elimination Cuts (SECs) for the TSP. However, adding linear
inequalities to the AP formulation changes the structure of the
relaxation which is no longer an AP. On the other hand, we are
interested in maintaining this structure since we have a
polynomial and incremental algorithm that solves the problem.
Therefore, as done in \cite{FLM2002}, we relax cuts in a
Lagrangean way, thus maintaining an AP structure.

The resulting relaxation, we call it AP$^{cuts}$, still has an AP
structure, but provides a tighter bound than the initial AP. More
precisely, it provides the same objective function value as the
linear relaxation where all cuts are added defining the sub-tour
polytope. In many cases, in particular for TSPTW instances, the
bound is extremely close to the optimal solution.

\subsection{Domain partitioning}
\label{ssc:partitioning} As described in Section \ref{problem},
the solution of an Assignment Problem provides the reduced cost
matrix with no additional computational cost. We recall that the
reduced cost $\overline{c}_{ij}$ of a variable $x_{ij}$
corresponds to the additional cost to be paid if this variable is
inserted in the solution, i.e., $x_{ij}=1$. Since these variables
are mapped into CP variables {\rm\it Next}, we obtain the same
esteem also for variable domain values. Thus, it is likely that
domain values that have a relative low reduced cost value will be
part of an optimal solution to the TSP.

This property is used to partition the domain $D_i$ of a variable
${\rm\it Next}_i$ into $D_i^{good}$ and $D_{i}^{bad}$, such that
$D_i = D_{i}^{good} \cup D_{i}^{bad} $ and $D_{i}^{good} \cap
D_{i}^{bad} = \emptyset$ for all $i=1,\dots, n$. Given a ratio $r
\geq 0 $, we define for each variable ${\rm\it Next}_i$ the good
set $D_i^{good}$ by selecting from the domain $D_i$ the values $j$
that have the $r*n$ lowest $\overline{c}_{ij}$. Consequently,
$D_i^{bad} = D_i \setminus D_i^{good}$.

The ratio defines the size of the good domains, and will be
discussed in Section~\ref{sc:quality}. Note that the {\em optimal}
ratio should be experimentally tuned, in order to obtain the
optimal solution in the first subproblem, and it is strongly
problem dependent. In particular, it depends on the structure of
the problem we are solving. For instance, for the pure TSP
instances considered in this paper, a good ratio is 0.05 or 0.075,
while for TSPTW the best ratio observed is around 0.15. With this
ratio, the optimal solution of the original problem is indeed
located in the first generated subproblem in almost all test
instances.

\subsection{LDS for generating subproblems}\label{ssc:generation}

In the previous section, we described how to partition variable
domains in a good and a bad set by exploiting information on
reduced costs. Now, we show how to explore a search tree on the
basis of this domain partitioning. At each node corresponding to
the choice of variable $X_i$, whose domain has been partitioned in
$D_i^{good}$ and $D_i^{bad}$, we impose on the left branch the
branching constraint $X_i \in D_i^{good}$, and on the right branch
$X_i \in D_i^{bad}$. Exploring with a Limited Discrepancy Search
strategy this tree, we first explore the subproblem suggested by
the heuristic where all variable range on the good set; then
subproblems where all variables but one range on the good set, and
so on.

If the reduced cost-based ranking criterion is accurate, as the
experimental results confirm, we are likely to find the optimal
solution in the subproblem $P^{(0)}$ generated by imposing all
variables ranging on the good set of values. If this heuristic
fails once, we are likely to find the optimal solution in one of
the $n$ subproblems ($P^{(1)}_i$ with $i \in \{1,\dots,n\}$)
generated by imposing all variables but one (variable $i$) ranging
on the good sets and one ranging on the bad set. Then, we go on
generating $n(n-1)/2$ problems $P^{(2)}_{ij}$ all variables but
two (namely $i$ and $j$) ranging on the good set and two ranging
on the bad set are considered, and so on.

In Section~\ref{implementation} we will see an implementation of
this search strategy that in a sense {\em squeezes} the subproblem
generation tree shown in Figure~\ref{fig:tree} into a constraint.

\subsection{Proof of optimality}

If we are simply interested in a good solution, without proving
optimality, we can stop our method after the solution of the first
generated subproblem. In this case, the proposed approach is
extremely effective since we almost always find the optimal
solution in that subproblem.

Otherwise, if we are interested in a provably optimal solution, we
have to prove optimality by solving all sub-problems at increasing
discrepancies. Clearly, even if all subproblems could be
efficiently solved, generating and solving all of them would not
be practical. However, if we exploit a tight initial lower bound,
as explained in Section~\ref{lagr}, which is successively improved
with considerations on the discrepancy, we can stop the generation
of subproblems after few trials since we prove optimality fast.

An important part of the proof of optimality is the management of
lower and upper bounds. The upper bound is decreased as we find
better solutions, and the lower bound is increased as a
consequence of discrepancy increase. The idea is to find an
optimal solution in the first subproblem, providing the best
possible upper bound.

The ideal case is that all subproblems but the first can be pruned
since they have a lower bound higher than the current upper bound,
in which case we only need to consider a single subproblem.

The initial lower bound ${\rm LB}_0$ provided by the Assignment
Problem AP$^{cuts}$ at the root node can be improved each time we
switch to a higher discrepancy $k$. For $i \in \{1, \dots, n\}$,
let $\overline{c}_i^*$ be the lowest reduced cost value associated
with $D_i^{bad}$, corresponding to the solution of AP$^{cuts}$,
i.e. $\overline{c}_i^* = \min_{j\in
D_i^{bad}}{\overline{c}_{ij}}$. Clearly, a first trivial bound is
${\rm LB}_0 + \min_{i \in \{1, \dots, n\}} \overline{c}_i^*$ for
all problems at discrepancy greater than or equal to 1.  We can
increase this bound: let $L$ be the nondecreasing ordered list of
$\overline{c}_i^*$ values, containing $n$ elements. $L[i]$ denotes
the $i$-th element in $L$. The following theorem achieves a better
bound improvement \cite{LodiPersonal}.

%


\begin{theorem} \label{th:bound}
For $k \in \{1,\dots, n\}$, ${\rm LB}_0 + \sum_{i=1}^{k} L[i]$ is
a valid lower bound for the subproblems corresponding to
discrepancy k.
\end{theorem}
\begin{proof}
The proof is based on the concept of additive bounding procedures
\cite{FT89,FT92} that states as follows: first we solve a
relaxation of a problem $P$. We obtain a bound $LB$, in our case
${\rm LB}_0$ and a reduced-cost matrix $\overline{c}$. Now we
define a second relaxation of $P$ having cost matrix
$\overline{c}$. We obtain a second lower bound $LB^{(1)}$. The sum
$LB + LB^{(1)}$ is a valid lower bound for $P$. In our case, the
second relaxation is defined by the constraints imposing in-degree
of each vertex less or equal to one plus a linear version of the
{\em k-discrepancy constraint}: $ \sum_{i \in V} \sum_{j \in
D_i^{bad}} x_{ij} = k $. Thus, $\sum_{i=1}^{k} L[i]$ is exactly
the optimal solution for this problem. \hfill $\Box$
\end{proof}

Note that in general reduced costs are not additive, but in this case they
are. As a consequence of this result, optimality is proven as soon as ${\rm
LB}_0 +
\sum_{i=1}^{k} L[i]
 > {\rm UB}$ for some discrepancy $k$, where ${\rm UB}$ is the current upper
bound.
We used this bound in our implementation.


\subsection{Solving each subproblem}\label{ssc:subpr}
Once the subproblems are generated, we can solve them with any complete
technique. In this paper,
we have used the method and the code described in \cite{NostroCP98} and in
\cite{FLM2002} for the
TSPTW. As a search heuristic we have used the one behaving best for each
problem.

\section{Implementation}
\label{implementation}

Finding a subproblem of discrepancy $k$ is equivalent to finding a set
$S~\subseteq~\{1,\dots,n\}$
with $|S| = k$ such that
\begin{displaymath}
\begin{array}{ll}
{\rm\it Next}_i \in D_i^{bad}  & $for $i \in S$, and $\\
{\rm\it Next}_i \in D_i^{good} & $for $i \in \{1, \dots, n\} \setminus S.
\end{array}
\end{displaymath}
The search for such a set $S$ and the corresponding domain assignments have
been
`squeezed' into a
constraint, the {\em discrepancy constraint} {\tt discr\_cst}. It takes as
input
the discrepancy
$k$, the variables ${\rm\it Next}_i$ and the domains $D_i^{good}$.
Declaratively, the constraint
holds if and only if exactly $k$ variables take their values in the bad sets.

Operationally, it keeps track of the number of variables that take their value
in either the good
or the bad domain. If during the search for a solution in the current
subproblem
the number of
variables ranging on their bad domain is $k$, all other variables are forced
to
range on their good
domain. Equivalently, if the number of variables ranging on their good domain
is
$n-k$, the other
variables are forced to range on their bad domain.

The subproblem generation is defined as follows (in pseudo-code):
\begin{verbatim}
   for (k=0..n) {
     add(discr_cst(k,next,D_good));
     solve subproblem;
     remove(discr_cst(k,next,D_good));
   }
\end{verbatim}

where {\tt k} is the level of discrepancy, {\tt next} is the array
containing ${\rm\it Next}_i$, and {\tt D\_good} is the array
containing $D_i^{good}$ for all $i \in \{1, \dots, n\}$. The
command {\tt solve subproblem} is shorthand for solving the
subproblem which has been considered in Section~\ref{ssc:subpr}.

A more traditional implementation of LDS (referred to as
`standard' in Table~\ref{tb:tree_vs_cst}) exploits tree search,
where at each node the domain of a variable is split into the good
set or the bad set, as described in Section~\ref{ssc:generation}.
In Table~\ref{tb:tree_vs_cst}, the performance of this traditional
approach is compared with the performance of the discrepancy
constraint (referred to as {\tt discr\_cst} in
Table~\ref{tb:tree_vs_cst}). In this table results on TSPTW
instances (taken from \cite{atsptw}) are reported. All problems
are solved to optimality and both approaches use a ratio of 0.15
to scale the size of the good domains. In the next section, this
choice is experimentally derived. Although one method does not
outperform the other, the overall performance of the discrepancy
constraint is in general slightly better than the traditional LDS
approach. In fact, for solving all instances, we have in the
traditional LDS approach a total time of 2.75  with 1465 fails,
while using the constraint we have 2.61 seconds and 1443 fails.

\begin{table}

\begin{minipage}[t]{0.49\textwidth}
\begin{tabular}{|p{1.5cm}|p{1.0cm}p{1.0cm}|p{1.0cm}p{1.0cm}|}
\hline \multicolumn{1}{|c|}{\mbox{}}  & \multicolumn{2}{|c|}{standard} &
\multicolumn{2}{|c|}{{\tt
discr\_cst}}\\ \hline instance & time & fails & time & fails \\ \hline
rbg016a & 0.08 & 44 & 0.04 & 44 \\
rbg016b & 0.15 & 57 & 0.10 & 43 \\
rbg017.2 & 0.05 & 14 & 0.05 & 14 \\
rbg017 & 0.10 & 69 & 0.09 & 47 \\
rbg017a & 0.09 & 42 & 0.09 & 42 \\
rbg019a & 0.06 & 30 & 0.06 & 30 \\
rbg019b & 0.12 & 71 & 0.12 & 71 \\
rbg019c & 0.20 & 152 & 0.19 & 158 \\
rbg019d & 0.06 & 6 & 0.06 & 6 \\
rbg020a & 0.07 & 5 & 0.06 & 5 \\ \hline
\end{tabular}
\end{minipage}
\hfil
\begin{minipage}[t]{0.49\textwidth}
\begin{tabular}{|p{1.5cm}|p{1.0cm}p{1.0cm}|p{1.0cm}p{1.0cm}|}
\hline \multicolumn{1}{|c|}{\mbox{}}  & \multicolumn{2}{|c|}{standard} &
\multicolumn{2}{|c|}{{\tt
discr\_cst}}\\ \hline instance & time & fails & time & fails \\ \hline
rbg021.2 & 0.13 & 66 & 0.13 & 66 \\
rbg021.3 & 0.22 & 191 & 0.20 & 158 \\
rbg021.4 & 0.11 & 85 & 0.09 & 40 \\
rbg021.5 & 0.10 & 45 & 0.16 & 125 \\
rbg021.6 & 0.19 & 110 & 0.2 & 110 \\
rbg021.7 & 0.23 & 70 & 0.22 & 70 \\
rbg021.8 & 0.15 & 88 & 0.15 & 88 \\
rbg021.9 & 0.17 & 108 & 0.17 & 108 \\
rbg021 & 0.19 & 152 & 0.19 & 158 \\
rbg027a & 0.22 & 53 & 0.21 & 53 \\ \hline
\end{tabular}
\end{minipage}
\newline
\caption{Comparison of traditional LDS and the discrepancy
constraint.}\label{tb:tree_vs_cst}
\end{table}

\section{Quality of Heuristic}\label{sc:quality}

In this section we evaluate the quality of the heuristic used. On
the one hand, we would like the optimal solution to be in the
first subproblem, corresponding to discrepancy 0. This subproblem
should be as small as possible, in order to be able to solve it
fast. On the other hand, we need to have a good bound to prove
optimality. For this we need relatively large reduced costs in the
bad domains, in order to apply Theorem~\ref{th:bound} effectively.
This would typically induce a larger first subproblem.
Consequently, we should make a tradeoff between finding a good
first solution and proving optimality. This is done by tuning the
ratio $r$, which determines the size of the first subproblem. We
recall from Section~\ref{ssc:partitioning} that $|D_i^{good}| \leq
rn$ for $i \in \{1,\dots,n\}$.

In Tables \ref{tb:ratio_tsp} and \ref{tb:ratio_tsptw} we report the quality
of the heuristic with respect to the ratio. The TSP instances are taken
from TSPLIB \cite{tsplib} and the asymmetric TSPTW instances are due to
Ascheuer \cite{atsptw}. All subproblems are solved to optimality with a
fixed strategy, as to make a fair comparison. In the tables, `size' is the
actual relative size of the first subproblem with respect to the initial
problem. The size is calculated by $\frac{1}{n}
\sum_{i=1}^n |D_i^{good}| / |D_i|$.

\begin{table}
\begin{tabular}{|l|rccr|rccr|rccr|rccr|}
\hline \multicolumn{1}{|c}{\mbox{}}  & \multicolumn{4}{c}{ratio=0.025} &
\multicolumn{4}{c}{ratio=0.05} & \multicolumn{4}{c}{ratio=0.075} &
\multicolumn{4}{c|}{ratio=0.1}
\\ \hline instance & size & opt & pr & fails & size & opt & pr &
fails & size & opt & pr & fails & size & opt & pr & fails \\
\hline
gr17     &  0.25    &0  &1  &2  &0.25   &0  &1  &2  &0.32   &0  &1  &7  &0.32
&0  &1  &7  \\
gr21     &  0.17    &0  &1  &1  &0.23   &0  &1  &6  &0.23   &0  &1  &6  &0.28
&0  &1  &14 \\
gr24     &  0.17    &0  &1  &2  &0.21   &0  &1  &2  &0.21   &0  &1  &2  &0.26
&0  &1  &39 \\
fri26    &  0.16    &0  &1  &1  &0.20    &0  &1  &4  &0.24   &0  &1  &327
&0.24   &0  &1  &327    \\
bayg29 &    0.14    &1  &2  &127    &0.17   &0  &1  &341    &0.21   &0  &1
&1k
&0.21   &0  &1  &1k \\
bays29 &    0.13    &1  &6  &19k    &0.16   &0  &2  &54 &0.20    &0  &1  &65
&0.20    &0  &1  &65\\
\hline
average & 0.17 & 0.33 & 2 & 22 & 0.20 & 0 & 1.17 & 68 & 0.24 & 0 & 1 & 68 &
0.25
& 0 & 1 & 75 \\
\hline
\end{tabular}
\newline
\caption{Quality of heuristic with respect to ratio for the
TSP.}\label{tb:ratio_tsp}
\end{table}

The domains in the TSPTW instances are typically much smaller than the number
of
variables, because
of the time window constraints that already remove a number of domain values.
Therefore, the first
subproblem might sometimes be relatively large, since only a few values are
left
after pruning, and
they might be equally promising.

The next columns in the tables are `opt' and `pr'. Here `opt'
denotes the level of discrepancy at which the optimal solution is
found. Typically, we would like this to be 0. The column `pr'
stands for the level of discrepancy at which optimality is proved.
This would preferably be 1. The column `fails' denotes the total
number of backtracks during search needed to solve the problem to
optimality.

\begin{table}
\begin{tabular}{|l|rccr|rccr|rccr|rccr|}
\hline
\multicolumn{1}{|c}{\mbox{}}  & \multicolumn{4}{c}{ratio=0.05} &
\multicolumn{4}{c}{ratio=0.1} & \multicolumn{4}{c}{ratio=0.15} &
\multicolumn{4}{c|}{ratio=0.2} \\ \hline
instance & size & opt & pr & fails & size & opt & pr & fails & size & opt &
pr &
fails & size & opt & pr & fails \\ \hline
rbg010a & 0.81 & 1 & 2 & 7 & 0.94 & 0 & 1 & 5 & 0.99 & 0 & 1 & 7 & 0.99 & 0 &
1
& 7 \\
rbg016a & 0.76 & 1 & 4 & 37 & 0.88 & 0 & 2 & 41 & 0.93 & 0 & 2 & 44 & 0.98 &
0 &
1 & 47 \\
rbg016b & 0.62 & 1 & 4 & 54 & 0.72 & 0 & 3 & 54 & 0.84 & 0 & 2 & 43 & 0.91 &
0 &
2 & 39 \\
rbg017.2 & 0.38 & 0 & 1 & 2 & 0.49 & 0 & 1 & 9 & 0.57 & 0 & 1 & 14 & 0.66 & 0
&
1 & 27 \\
rbg017 & 0.53 & 1 & 9 & 112 & 0.66 & 1 & 5 & 70 & 0.79 & 0 & 3 & 47 & 0.89 &
0 &
2 & 49 \\
rbg017a & 0.64 & 0 & 1 & 10 & 0.72 & 0 & 1 & 13 & 0.80 & 0 & 1 & 42 & 0.88 &
0 &
1 & 122 \\
rbg019a & 0.89 & 0 & 1 & 17 & 0.98 & 0 & 1 & 27 & 1 & 0 & 1 & 30 & 1 & 0 & 1 &
30 \\
rbg019b & 0.68 & 1 & 3 & 83 & 0.78 & 1 & 3 & 85 & 0.87 & 0 & 1 & 71 & 0.95 &
0 &
1 & 71 \\
rbg019c & 0.52 & 1 & 4 & 186 & 0.60 & 1 & 3 & 137 & 0.68 & 1 & 2 & 158 & 0.76
&
0 & 2 & 99 \\
rbg019d & 0.91 & 0 & 2 & 5 & 0.97 & 0 & 1 & 6 & 1 & 0 & 1 & 6 & 1 & 0 & 1 & 6
\\
rbg020a & 0.78 & 0 & 1 & 3 & 0.84 & 0 & 1 & 3 & 0.89 & 0 & 1 & 5 & 0.94 & 0 &
1
& 5 \\
rbg021.2 & 0.50 & 0 & 2 & 23 & 0.59 & 0 & 1 & 48 & 0.66 & 0 & 1 & 66 & 0.74 &
0
& 1 & 148 \\
rbg021.3 & 0.48 & 1 & 5 & 106 & 0.55 & 1 & 4 & 185 & 0.62 & 0 & 3 & 158 &
0.68 &
0 & 3 & 163 \\
rbg021.4 & 0.47 & 1 & 4 & 152 & 0.53 & 0 & 3 & 33 & 0.61 & 0 & 3 & 40 & 0.67
& 0
& 2 & 87 \\
rbg021.5 & 0.49 & 1 & 3 & 160 & 0.55 & 0 & 2 & 103 & 0.61 & 0 & 2 & 125 &
0.67 &
0 & 2 & 206 \\
rbg021.6 & 0.41 & 1 & 2 & 8k & 0.48 & 1 & 2 & 233 & 0.59 & 0 & 2 & 110 & 0.67
&
0 & 1 & 124 \\
rbg021.7 & 0.40 & 1 & 3 & 518 & 0.49 & 0 & 2 & 91 & 0.59 & 0 & 1 & 70 & 0.65
& 0
& 1 & 70 \\
rbg021.8 & 0.39 & 1 & 3 & 13k & 0.48 & 1 & 2 & 518 & 0.56 & 0 & 1 & 88 & 0.63
&
0 & 1 & 88 \\
rbg021.9 & 0.39 & 1 & 3 & 13k & 0.48 & 1 & 2 & 574 & 0.56 & 0 & 1 & 108 &
0.63 &
0 & 1 & 108 \\
rbg021 & 0.52 & 1 & 4 & 186 & 0.60 & 1 & 3 & 137 & 0.68 & 1 & 2 & 158 & 0.76
& 0
& 2 & 99 \\
rbg027a & 0.44 & 0 & 3 & 15 & 0.59 & 0 & 2 & 35 & 0.66 & 0 & 1 & 53 & 0.77 &
0 &
1 & 96 \\
\hline average & 0.57 & 0.67 & 3.05 & 1725 & 0.66 & 0.38 & 2.14 & 114 & 0.74 &
0.10 & 1.57 &
69 & 0.80 & 0 & 1.38 & 81 \\
\hline
\end{tabular}

\vspace{0.5cm}

\begin{tabular}{|l|rccr|rccr|rccr|}
\hline
\multicolumn{1}{|c}{\mbox{}} & \multicolumn{4}{c}{ratio=0.25} &
\multicolumn{4}{c}{ratio=0.3} & \multicolumn{4}{c|}{ratio=0.35} \\ \hline
instance & size & opt & pr & fails & size & opt & pr & fails & size & opt &
pr &
fails \\ \hline
rbg010a & 1 & 0 & 1 & 8 & 1 & 0 & 1 & 8 & 1 & 0 & 1 & 8 \\
rbg016a & 1 & 0 & 1 & 49 & 1 & 0 & 1 & 49 & 1 & 0 & 1 & 49 \\
rbg016b & 0.97 & 0 & 1 & 36 & 0.99 & 0 & 1 & 38 & 1 & 0 & 1 & 38 \\
rbg017.2 & 0.74 & 0 & 1 & 31 & 0.84 & 0 & 1 & 20 & 0.90 & 0 & 1 & 20 \\
rbg017 & 0.95 & 0 & 2 & 58 & 0.99 & 0 & 1 & 56 & 1 & 0 & 1 & 56 \\
rbg017a & 0.93 & 0 & 1 & 143 & 0.97 & 0 & 1 & 165 & 1 & 0 & 1 & 236 \\
rbg019a & 1 & 0 & 1 & 30 & 1 & 0 & 1 & 30 & 1 & 0 & 1 & 30 \\
rbg019b & 0.97 & 0 & 1 & 72 & 1 & 0 & 1 & 72 & 1 & 0 & 1 & 72 \\
rbg019c & 0.82 & 0 & 2 & 106 & 0.92 & 0 & 1 & 112 & 0.95 & 0 & 1 & 119 \\
rbg019d & 1 & 0 & 1 & 6 & 1 & 0 & 1 & 6 & 1 & 0 & 1 & 6 \\
rbg020a & 0.99 & 0 & 1 & 5 & 1 & 0 & 1 & 5 & 1 & 0 & 1 & 5 \\
rbg021.2 & 0.80 & 0 & 1 & 185 & 0.91 & 0 & 1 & 227 & 0.95 & 0 & 1 & 240 \\
rbg021.3 & 0.78 & 0 & 2 & 128 & 0.90 & 0 & 2 & 142 & 0.94 & 0 & 1 & 129 \\
rbg021.4 & 0.74 & 0 & 2 & 114 & 0.88 & 0 & 1 & 67 & 0.94 & 0 & 1 & 70 \\
rbg021.5 & 0.74 & 0 & 1 & 193 & 0.88 & 0 & 1 & 172 & 0.94 & 0 & 1 & 173 \\
rbg021.6 & 0.73 & 0 & 1 & 132 & 0.89 & 0 & 1 & 145 & 0.95 & 0 & 1 & 148 \\
rbg021.7 & 0.71 & 0 & 1 & 71 & 0.82 & 0 & 1 & 78 & 0.89 & 0 & 1 & 78 \\
rbg021.8 & 0.71 & 0 & 1 & 89 & 0.83 & 0 & 1 & 91 & 0.90 & 0 & 1 & 95 \\
rbg021.9 & 0.71 & 0 & 1 & 115 & 0.83 & 0 & 1 & 121 & 0.90 & 0 & 1 & 122 \\
rbg021 & 0.82 & 0 & 2 & 106 & 0.92 & 0 & 1 & 112 & 0.95 & 0 & 1 & 119 \\
rbg027a & 0.83 & 0 & 1 & 245 & 0.90 & 0 & 1 & 407 & 0.96 & 0 & 1 & 1k \\
\hline
average & 0.85 & 0 & 1.24 & 92 & 0.93 & 0 & 1.05 & 101 & 0.96 & 0 & 1 & 139 \\
\hline
\end{tabular}
\newline
  \caption{Quality of heuristic with respect to ratio for the asymmetric
TSPTW.}\label{tb:ratio_tsptw}
\end{table}

Concerning the TSP instances, already for a ratio of $0.05$ all
solutions are in the first subproblem. For a ratio of $0.075$, we
can also prove optimality at discrepancy 1. Taking into account
also the number of fails, we can argue that both $0.05$ and
$0.075$ are good ratio candidates for the TSP instances.

For the TSPTW instances, we notice that a smaller ratio does not
necessarily increase the total number of fails, although it is
more difficult to prove optimality. Hence, we have a slight
preference for the ratio to be $0.15$, mainly because of its
overall (average) performance.

An important aspect we are currently investigating is the dynamic tuning of
the ratio.

\section{Computational Results}\label{sc:results}

We have implemented and tested the proposed method using ILOG
Solver and Scheduler \cite{ilogsolver,ilogsched}. The algorithm
runs on a Pentium 1Ghz, 256 MB RAM, and uses CPLEX 6.5 as LP
solver. The two sets of test instances are taken from the TSPLIB
\cite{tsplib} and Ascheuer's asymmetric TSPTW problem instances
\cite{atsptw}.

\begin{table}
\begin{center}
\begin{tabular}{|l|p{1.0cm}p{1.0cm}|p{1.0cm}p{1.0cm}p{1.0cm}|p{0.8cm}p{1.0cm}p
{1.0cm}|}
\hline \multicolumn{1}{|c}{\mbox{}} & \multicolumn{2}{|c|}{No LDS} &
\multicolumn{3}{|c|}{LDS} &
\multicolumn{3}{|c|}{first subproblem}\\ \hline instance & time & fails &
time &
fails & ratio &
obj & time & fails \\ \hline
gr17 & 0.12 & 34 & 0.12 & 1 & 0.05 & - & - & - \\
gr21 & 0.07 & 19 & 0.06 & 1 & 0.05 & - & - & -\\
gr24 & 0.18 & 29 & 0.18 & 1 & 0.05 &-&-&-\\
fri26 & 0.17 & 70 & 0.17 & 4 & 0.05 &-&-&-\\
bayg29 & 0.33 & 102 & 0.28 & 28 & 0.05 &-&-&-\\
bays29$^*$ & 0.30 & 418 & 0.20 & 54 & 0.05 & opt & 0.19 & 36 \\
dantzig42 & 1.43 & 524 & 1.21 & 366 & 0.075&-&-&- \\
hk48 & 12.61 & 15k & 1.91 & 300 & 0.05&-&-&- \\
gr48$^*$ & 21.05 & 25k & 19.10 & 22k & 0.075 & opt & 5.62 & 5.7k \\
brazil58$^*$ & limit & limit & 81.19 & 156k & 0.05 & opt & 80.3 & 155k \\
\hline
\multicolumn{9}{l}{$^*$ Optimality not proven directly after solving first
subproblem.}
\end{tabular}
  \newline
   \caption{Computational results for the TSP.}\label{tb:results_tsp}
\end{center}
\end{table}
\begin{table}
\begin{minipage}[t]{0.48\textwidth}
\begin{tabular}{|l|p{0.65cm}p{0.65cm}|p{0.65cm}p{0.65cm}|p{0.65cm}p{0.65cm}|}
\hline
\multicolumn{1}{|c}{\mbox{}} & \multicolumn{2}{|c|}{FLM2002} &
\multicolumn{2}{|c|}{No LDS} &
\multicolumn{2}{|c|}{LDS} \\ \hline instance & time & fails & time & fails &
time & fails \\ \hline
rbg010a & 0.0 & 6 & 0.06 & 6 & 0.03 & 4 \\
rbg016a & 0.1 & 21 & 0.04 & 10 & 0.04 & 8 \\
rbg016b & 0.1 & 27 & 0.11 & 27 & 0.09 & 32 \\
rbg017.2 & 0.0 & 17 & 0.04 & 14 & 0.04 & 2 \\
rbg017 & 0.1 & 27 & 0.06 & 9 & 0.08 & 11 \\
rbg017a & 0.1 & 22 & 0.05 & 8 & 0.05 & 1 \\
rbg019a & 0.0 & 14 & 0.05 & 11 & 0.05 & 2 \\
rbg019b & 0.2 & 80 & 0.08 & 37 & 0.09 & 22 \\
rbg019c & 0.3 & 81 & 0.08 & 17 & 0.14 & 74 \\
rbg019d & 0.0 & 32 & 0.06 & 19 & 0.07 & 4 \\
rbg020a & 0.0 & 9 & 0.07 & 11 & 0.06 & 3 \\
rbg021.2 & 0.2 & 44 & 0.09 & 20 & 0.08 & 15 \\
rbg021.3 & 0.4 & 107 & 0.11 & 52 & 0.14 & 80 \\
rbg021.4 & 0.3 & 121 & 0.10 & 48 & 0.09 & 32 \\
rbg021.5 & 0.2 & 55 & 0.14 & 89 & 0.12 & 60 \\
rbg021.6 & 0.7 & 318 & 0.14 & 62 & 0.16 & 50 \\
\hline
\end{tabular}
\end{minipage}
\begin{minipage}[t]{0.50\textwidth}
\begin{tabular}{|l|p{0.725cm}p{0.65cm}|p{0.725cm}p{0.6cm}|p{0.725cm}p{0.65cm}
|}
\hline
\multicolumn{1}{|c}{\mbox{}} & \multicolumn{2}{|c|}{FLM2002} &
\multicolumn{2}{|c|}{No LDS} &
\multicolumn{2}{|c|}{LDS} \\ \hline instance & time & fails & time & fails &
time & fails \\ \hline
rbg021.7 & 0.6 & 237 & 0.19 & 45 & 0.21 & 43 \\
rbg021.8 & 0.6 & 222 & 0.09 & 30 & 0.10 & 27 \\
rbg021.9 & 0.8 & 310 & 0.10 & 31 & 0.11 & 28 \\
rbg021 & 0.3 & 81 & 0.07 & 17 & 0.14 & 74 \\
rbg027a & 0.2 & 50 & 0.19 & 45 & 0.16 & 23 \\
rbg031a & 2.7 & 841 & 0.58 & 121 & 0.68 & 119 \\
rbg033a & 1.0 & 480 & 0.62 & 70 & 0.73 & 55 \\
rbg034a & 55.2 & 13k & 0.65 & 36 & 0.93 & 36 \\
rbg035a.2 & 36.8 & 5k & 5.23 & 2.6k & 8.18 & 4k \\
rbg035a & 3.5 & 841 & 0.82 & 202 & 0.83 & 56 \\
rbg038a & 0.2 & 49 & 0.37 & 42 & 0.36 & 3 \\
rbg040a & 738.1 & 136k & 185.0 & 68k & 1k & 387k \\
rbg042a & 149.8 & 19k & 29.36 & 11k & 70.71 & 24k \\
rbg050a & 180.4 & 19k & 3.89 & 1.6k & 4.21 & 1.5k \\
rbg055a & 2.5 & 384 & 4.39 & 163 & 4.50 & 133 \\
rbg067a & 4.0 & 493 & 26.29 & 171 & 25.69 & 128 \\
\hline
\end{tabular}
\end{minipage}
\newline
\caption{Computational results for the asymmetric
TSPTW.}\label{tb:results_tsptw}
\end{table}
\begin{table}
\begin{minipage}[t]{0.49\textwidth}
\begin{tabular}{|p{1.5cm}|p{0.65cm}p{0.65cm}p{0.7cm}|p{0.8cm}p{0.8cm}|}
\hline instance & opt & obj & gap (\%) & time & fails \\ \hline
rbg016a & 179 & 179 & 0 & 0.04 & 7 \\
rbg016b & 142 & 142 & 0 & 0.09 & 22\\
rbg017 & 148 & 148 & 0 & 0.07 & 5\\
rbg019c & 190 & 202 & 6.3 & 0.11 & 44\\
rbg021.3 & 182 & 182 & 0 & 0.09 & 29\\
rbg021.4 & 179 & 179 & 0 & 0.06 & 4\\ \hline
\end{tabular}
\end{minipage}
\hfil
\begin{minipage}[t]{0.49\textwidth}
\begin{tabular}{|p{1.5cm}|p{0.65cm}p{0.65cm}p{0.7cm}|p{0.85cm}p{0.85cm}|}
\hline instance & opt & obj & gap (\%) & time & fails \\ \hline
rbg021.5 & 169 & 169 & 0 & 0.10 & 19\\
rbg021.6 & 134 & 134 & 0 & 0.15 & 49\\
rbg021 & 190 & 202 & 6.3 & 0.10 & 44\\
rbg035a.2 & 166 & 166 & 0 & 4.75 & 2.2k\\
rgb040a & 386 & 386 & 0 & 65.83 & 25k\\
rbg042a & 411 & 411 & 0 & 33.26 & 11.6k\\ \hline
\end{tabular}
\end{minipage}
\newline
\caption{Computational results for the first subproblem of the asymmetric
TSPTW}\label{tb:lds_0}
\end{table}

Table \ref{tb:results_tsp} shows the results for small TSP
instances. Time is measured in seconds, fails again denote the
total number of backtracks to prove optimality. The time limit is
set to 300 seconds. Observe that our method (LDS) needs less
number of fails than the approach without subproblem generation
(No LDS). This comes with a cost, but still our approach is never
slower and in some cases considerably faster. The problems were
solved both with a ratio of 0.05 and 0.075, the best of which is
reported in the table. Observe that in some cases a ratio of 0.05
is best, while in other cases 0.075 is better. For three instances
optimality could not be proven directly after solving the first
subproblem. Nevertheless the optimum was found in this subproblem
(indicated by objective `obj' is `opt'). Time and the number of
backtracks (fails) needed in the first subproblem are reported for
these instances.

In Table~\ref{tb:results_tsptw} the results for the asymmetric TSPTW instances
are shown. Our
method (LDS) uses a ratio of 0.15 to solve all these problems to optimality.
It
is compared to our
code without the subproblem generation (No LDS), and to the results by
Focacci,
Lodi and Milano
(FLM2002) \cite{FLM2002}. Up to now, FLM2002 has the fastest solution times
for
this set of
instances, to our knowledge. When comparing the time results (measured in
seconds), one should take
into account that FLM2002 uses a Pentium III 700 MHz.

Our method behaves in general quite well. In many cases it is much faster than
FLM2002. However, in
some cases the subproblem generation does not pay off. This is for instance
the
case for {\tt
rbg040a} and {\tt rbg042a}. Although our method finds the optimal solution in
the first branch
(discrepancy 0) quite fast, the initial bound LB$_0$ is too low to be able to
prune the search tree
at discrepancy 1. In those cases we need more time to prove optimality than we
would have needed if
we did not apply our method (No LDS). In such cases our method can be applied
as
an effective {\em
incomplete} method, by only solving first subproblem. Table~\ref{tb:lds_0}
shows
the results for
those instances for which optimality could not be proven directly after
solving
the first
subproblem. In almost all cases the optimum is found in the first subproblem.

\section{Discussion and Conclusion}\label{sc:discussion}

We have introduced an effective search procedure that consists
of generating promising subproblems and solving them.
To generate the subproblems, we split the variable domains in a
good part and a bad part on the basis of reduced costs. The domain
values corresponding to the lowest reduced costs are more likely
to be in the optimal solution, and are put into the good set.

The subproblems are generated using a LDS strategy, where the discrepancy is
the number of variables ranging on their bad set. Subproblems are
considerably smaller than
the original problem, and can be solved faster. To prove optimality, we
introduced a way of
increasing the lower bound using information from the discrepancies.

Computational results on TSP and asymmetric TSPTW instances show
that the proposed ranking is extremely accurate. In almost all cases
the optimal solution is found in the first subproblem. When
proving optimality is difficult, our method can still be used as
an effective incomplete search procedure, by only solving the
first subproblem.

Some interesting points arise from the paper: first, we have seen
that reduced costs represent a good ranking criterion for variable
domain values. The ranking quality is not affected if reduced
costs come from a loose relaxation.

Second, the tightness of the lower bound is instead very important
for the proof of optimality. Therefore, we have
used a tight bound at the root node and increased it thus
obtaining a discrepancy-based bound.

Third, if we are not interested in a complete algorithm, but we
need very good solutions fast, our method turns out to be a very
effective choice, since the first generated subproblem almost
always contains the optimal solution.

Future directions will explore a different way of generating
subproblems. Our domain partitioning is statically defined only at the root
node and maintained during the search. However, although being
static, the method is still very effective. We will explore a
dynamic subproblem generation.

\section*{Acknowledgements}
We would like to thank Filippo Focacci and Andrea Lodi for useful
discussion, suggestions and ideas.


\end{document}